\documentclass{article}

\usepackage{microtype}
\usepackage{graphicx}
\usepackage{subfigure}
\usepackage{booktabs} 
\usepackage[normalem]{ulem}
\useunder{\uline}{\ul}{}
\usepackage{graphicx}
\usepackage[table,xcdraw]{xcolor}
\usepackage{xcolor}
\usepackage{ulem}
\usepackage{hyperref}

\usepackage[accepted]{icml2024}

\usepackage{amsmath}
\usepackage{amssymb}
\usepackage{mathtools}
\usepackage{amsthm}
\usepackage{makecell}
\usepackage{bbding}
\usepackage{float}

\usepackage[capitalize,noabbrev]{cleveref}

\theoremstyle{plain}

\theoremstyle{definition}

\theoremstyle{remark}

\usepackage[textsize=tiny]{todonotes}

\icmltitlerunning{DenseMamba: State Space Models with Dense Hidden Connection for Efficient Large Language Models}

\begin{document}

\twocolumn[
\icmltitle{DenseMamba: State Space Models with Dense Hidden Connection\\for Efficient Large Language Models}



\icmlsetsymbol{equal}{*}

\begin{icmlauthorlist}
\icmlauthor{Wei He}{equal,yyy}
\icmlauthor{Kai Han}{equal,yyy}
\icmlauthor{Yehui Tang}{yyy}
\icmlauthor{Chengcheng Wang}{yyy}
\icmlauthor{Yujie Yang}{yyy}
\icmlauthor{Tianyu Guo}{yyy}
\icmlauthor{Yunhe Wang}{yyy}

\end{icmlauthorlist}

\icmlaffiliation{yyy}{Huawei Noah’s Ark Lab}

\icmlcorrespondingauthor{Kai Han}{kai.han@huawei.com}
\icmlcorrespondingauthor{Yunhe Wang}{yunhe.wang@huawei.com}

\icmlkeywords{Machine Learning, ICML}

\vskip 0.3in
]



\printAffiliationsAndNotice{\icmlEqualContribution} 

\begin{abstract}
	Large language models (LLMs) face a daunting challenge due to the excessive computational and memory requirements of the commonly used Transformer architecture. While state space model (SSM) is a new type of foundational network architecture offering lower computational complexity, their performance has yet to fully rival that of Transformers. This paper introduces DenseSSM, a novel approach to enhance the flow of hidden information between layers in SSMs. By selectively integrating shallow-layer hidden states into deeper layers, DenseSSM retains fine-grained information crucial for the final output. Dense connections enhanced DenseSSM still maintains the training parallelizability and inference efficiency. The proposed method can be widely applicable to various SSM types like RetNet and Mamba. With similar model size, DenseSSM achieves significant improvements, exemplified by DenseRetNet outperforming the original RetNet with up to 5\% accuracy improvement on public benchmarks. code is avalaible at :~\url{https://github.com/WailordHe/DenseSSM} .
\end{abstract}

\section{Introduction}
\label{sec:intro}

Since the release of ChatGPT~\cite{openai2023chatgpt}, large language models have entered a new epoch, showcasing outstanding abilities in language comprehension, dialogue, and logical reasoning. Over the past year, the industry has witnessed the emergence of numerous large language models, such as LLaMA~\cite{touvron2023llama} and ChatGLM~\cite{zeng2023glm-130b}. These large language models have given rise to a plethora of practical applications, including conversational bots, code assistants, and AI agents. The foundation of large language models lies in the Transformer network structure~\cite{vaswani2017attention}, primarily utilizing a multi-head self-attention module for modeling relationships between tokens and a Feed-forward network for non-linear feature transformations. The scaling law~\cite{kaplan2020scaling} based on the Transformer structure has propelled the continuous development and expansion of large language models.

In the Transformer network, multi-head self-attention (MHSA) plays a crucial role, but it comes with significant computational demands and memory requirements during inference. In terms of computational complexity, for an input sentence of length $N$, the calculation of self-attention has a complexity of $O(N^2)$ during training and inference. Regarding memory usage, previously encountered keys and values are stored, leading to a memory occupation of $O(ND)$.
As a result, recent efforts on network architectures have focused on simplifying Transformer by reducing its computation and space complexity. This includes various approaches, notably convolutional language models~\cite{poli2023hyena}, recurrent unit~\cite{lei2021attention}, long context models~\cite{ding2023longnet}, and state space models (SSMs)~\cite{gu2021efficiently,mamba}. These new models have provided strong alternatives to Transformer for building efficient LLMs. 

SSMs propose modeling sequences by introducing an appropriate design of hidden states for handling long-range dependencies with both training parallelizability and inference efficiency. Starting from the continuous mapping system, SSMs are discretized to process discrete inputs in deep learning such as language sequence. The discretized SSMs can be computed in both linear recurrence and global convolution modes. Commonly, convolution mode is used during training to achieve parallel acceleration, while recurrence mode is used during autoregressive inference because it has lower computational complexity.

The core distinction of SSMs from other neural networks, such as fully-connected neural networks, lies in the design of hidden states. Hidden states enable information to be propagated along the temporal dimension, while avoiding the computation complexity of accessing historical tokens at each step. Through state transition parameters $A$, hidden states transfer the hidden information from the previous time steps to the current time step, allowing for autoregressive prediction of the next token. Hidden states play a crucial role in SSMs, but have not received sufficient investigation in the past. Weights and hidden features in different layers contain information at various levels from fine-grained to coarse-grained~\cite{gu2021efficiently}. However, in previous versions of SSMs, hidden states only flowed within the current layer and could not transmit more information to deeper layers, thus failing to capture more hierarchical information.

In this paper, we propose DenseSSM to facilitate a more comprehensive flow of hidden information between layers in state space models. We first analyze the hidden state degradation in conventional SSMs which will prevent hidden information flow from low levels to high levels. By selectively integrating shallow-layer hidden states into deeper layers, DenseSSM retains fine-grained information that is useful for the final output. The proposed method is applicable to different types of SSMs, such as RetNet~\cite{retnet} and Mamba~\cite{mamba}. Our approach maintains the training parallelizability and inference efficiency of SSMs, while achieving a significant improvement with only a slight increase in the number of parameters. For instance, our DenseRetNet model outperforms traditional RetNet with up to 5\% accuracy improvement on public benchmarks.

\section{Related Works}

\subsection{Large Language Models}
Large language models (LLMs) have seen transformative advancements, enabling them to excel in a diverse array of natural language processing (NLP) tasks, including machine translation, text summarization, and emergent abilities like incontext learning, which were previously unattainable by earlier language models~\cite{devlin2019bert, raffel2023exploring}. The evolution of LLMs has been marked by a monumental shift in scale, exemplified by models like GPT-3~\cite{brown2020language}, with its 175 billion parameters, and the even more expansive PaLM~\cite{chowdhery2022palm}, packing in a astounding 540 billion parameters. These models have empirically validated the scaling law~\cite{kaplan2020scaling}, which posits that increasing model size leads to improved performance.

The rapid expansion in model size has underscored the critical need for the development of efficient Transformer algorithms, where FlashAttention~\cite{dao2022flashattention, dao2023flashattention2} has emerged as a significant innovation. This approach enhances the pivotal attention mechanism within Transformers by optimizing softmax computations using a technique known as tiling. By minimizing memory transactions between the GPU's HBM and on-chip SRAM, FlashAttention compute exact attention with fewer memory accesses, resulting in both faster execution and a lower memory footprint compared to standard attention implementations.

\subsection{State Space Models}
While the Transformer is currently the de facto architecture for large language models (LLMs), providing efficient parallel GPU training, the inference time for single-token inference increases significantly with longer sequence lengths, posing challenges for deployment due to the O(N) complexity per step even with accelerating algorithms like FlashAttention~\cite{dao2022flashattention, dao2023flashattention2}. 
Efforts have been dedicated to researching the Transformer-Next architecture, aiming to achieve state-of-the-art (SOTA) performance with efficient parallel training and effective inference, particularly for long sequence lengths.

State Space Sequence Models (SSMs) have recently emerged as promising architectures for sequence modeling. HiPPO~\cite{gu2020hippo} streamlines sequence modeling by compressing lengthy inputs into a dynamic, polynomial-based representation using orthogonal polynomials. S4~\cite{gu2021efficiently} introduced a novel parameterization through the application of a low-rank structured correction, enabling stable diagonalization and simplifying the process into Cauchy kernel operations. S5~\cite{smith2023simplified} further simplifies the S4 layer by employing a single multi-input, multi-output SSM and introducing efficient parallel scan algorithms into the S4 layers. H3~\cite{fu2023hungry} narrows the performance gap between SSMs and Transformer language models by designing three projections (Q, K, V) to simulate the attention mechanism and adopting a fast Fourier transform (FFT) to reduce computation and memory consumption further.

GSS~\cite{mehta2022long} was the first gated neural network architecture incorporating SSMs, it builds upon ~\cite{hua2022transformer} and introducing a compact SSM architecture that contracts model dimensions.
Unlike GSS, which emphasizes compressing context into a smaller state, Mamba~\cite{mamba} diverges by focusing on enhancing the selectivity of the state representation, aiming to balance the tradeoff between efficiency and effectiveness without compromising the model's ability to capture essential information from the context. It achieves this by integrating a selection mechanism which enabling the model to selectively prioritize relevant information while concurrently utilizing a hardware-optimized algorithm that ensures efficient computation.

\subsection{Linear Attention}
Linear attentions~\cite{katharopoulos2020transformers, zhai2021attention}, which remove the softmax operation from traditional attention, can be seen as a derivative of State Space Models (SSMs). They replace SSMs' convolutions with a variation of Multi-Head Attention (MHA) and eliminate the softmax of the traditional attention mechanism by utilizing a kernel function that operates independently on the queries (Q) and keys (K). These mechanisms also have a parallel form for efficient training and a recurrent form with $O(1)$ complexity.

RetNet~\cite{retnet}, TransNormerLLM~\cite{qin2024transnormerllm}, and RWKV~\cite{rwkv} implement a fixed decay factor to update the previous key-value (KV) states at each recurrent step. This decay mechanism seamlessly integrates with the causal attention mask for efficient parallel computation. However, since this decay factor is preset and independent of the data, it may not be universally applicable across all tasks, especially when prompts or long-range information is particularly important. 
To address this challenge, GLA (Gated Linear Attention) ~\cite{yang2023gated} introduces data-dependent gating mechanisms that are practical for both parallel and block-parallel forms. It performs competitively against strong baselines, including the LLaMA-architecture Transformer~\cite{touvron2023llama} and Mamba~\cite{mamba}.

\section{DenseSSM}
\label{sec:method}
In this section, we analyze the hidden state degradation in the deeper layers of SSMs and further introduce dense connection of hidden states to preserve richer information for deeper layers.

\subsection{Prelimineries}
\paragraph{Transformer}
Transformer is the widely-used network architecture of large language models which is based on the self-attention mechanism. The self-attention performs as follows:
\begin{equation}
	o_t = W_o\frac{\sum_{i=1}^Te^{q_t^Tk_i}v_i}{\sum_{i=1}^Te^{q_t^Tk_i}}l,
\end{equation}
where $q$, $k$ and $v$ are obtained by fully-connected layers, $W_o$ is the linear transformation weight for the output token $o_t$ at the $t$-th timestep. Each token will merge information of the other tokens by relationship weights calculated by the self-attention. In addition to self-attention module, the fee-forward network (FFN) module is another key component to transform the token representation and introduces more non-linearity. FFN module is usually composed by two stacked linear layers and non-linear activation function:
\begin{equation}
    y_t = W_{down}\sigma(W_{up}o_t),
\end{equation}
where $W_{up}$ and $W_{down}$ are the weight matrices of up projection and down projection layers, and $\sigma(\cdot)$ is the activation function such as GELU~\cite{hendrycks2016gaussian}.

\begin{figure*}[htp]
	\centering
	\small
	\setlength{\tabcolsep}{10pt}{
		\begin{tabular}{cc}
			\makecell*[c]{\includegraphics[width=0.48\linewidth]{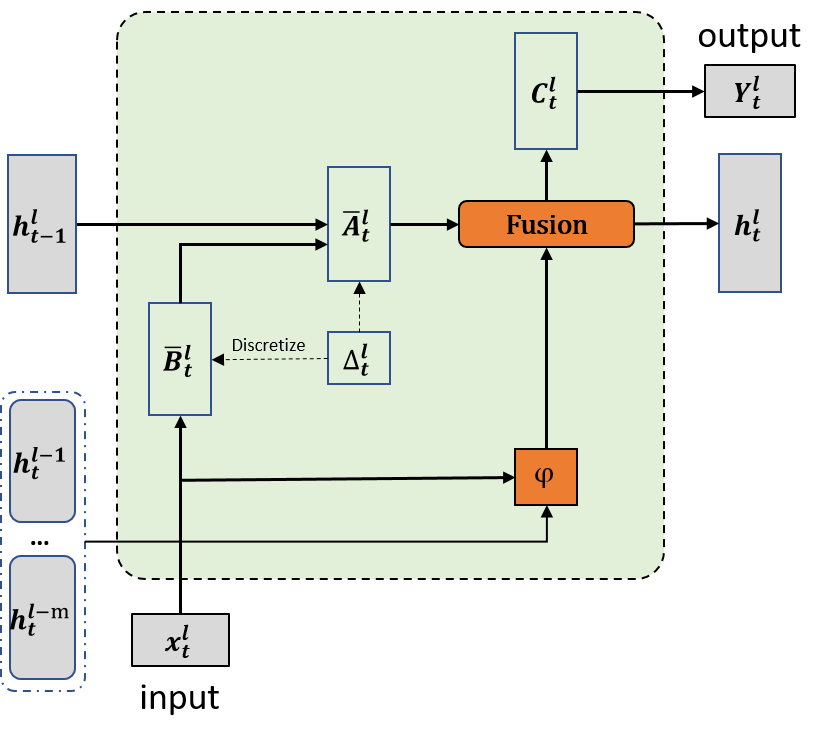}}  &
			\makecell*[c]{\includegraphics[width=0.43\linewidth]{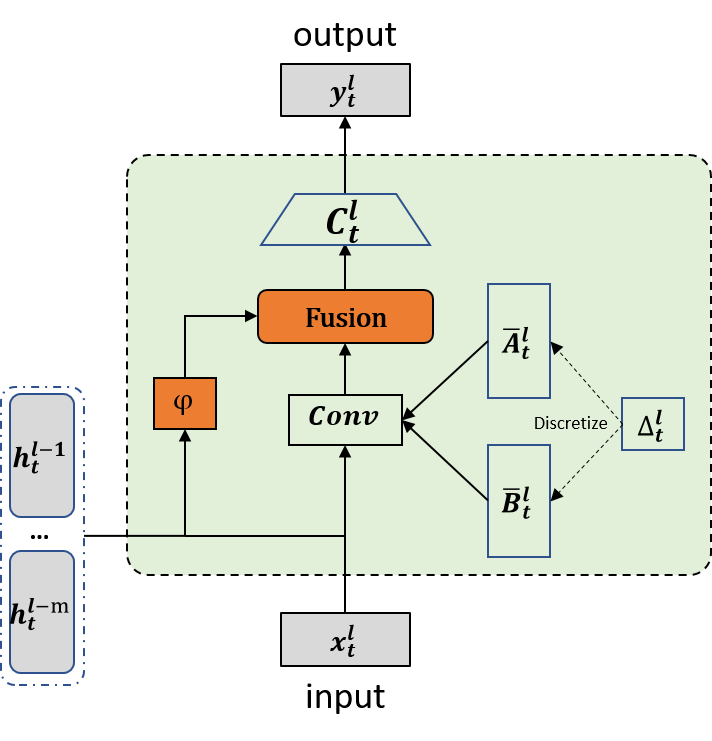}}  
			
			\\
			(a) DenseSSM in autoregressive mode. & (b) DenseSSM in parallelizable convolution mode.
		\end{tabular}
	}
	\caption{Illustrations of DenseSSM framework, where $\phi$ is the selective transition module and `Fusion' is the hidden fusion module.}
	\label{Fig:dense_ssm}
\end{figure*}

\paragraph{SSM}
State space models (SSM) in the literature of deep learning refer to the class of structured SSMs~\cite{gu2021efficiently} and the derivatives such as RWKV~\cite{rwkv} and RetNet~\cite{retnet}. Here we briefly describe the structured SSMs as a representative. Structured SSMs define a sequence-to-sequence transformation $x(t)\rightarrow y(t)$ with an implicit latent state $h(t)$. The continuous form is formulated as
\begin{align}
	h'(t) &= Ah(t) + Bx(t),\\
	y(t) &= Ch(t),
\end{align}
where $A$, $B$ and $C$ are the parameters. To apply SSM to the real discrete data, we discretize the continuous case and obtain the recurrence formulation and convolution formulation of it. The parameters $A$ and $B$ are transformed to the discrete parameters $\overline{A}$ and $\overline{B}$ with the discretization rule such as zero-order hold~\cite{gu2021efficiently}. The recurrence formulation is
\begin{align}
	h_t &= \overline{A}h_{t-1} + \overline{B}x_t,\label{eq:recurrence1}\\
	y_t &= Ch_t.\label{eq:recurrence}
\end{align}
The convolution formulation is
\begin{align}\label{eq:conv}
	\overline{K} &= (C\overline{B},C\overline{AB},\cdots,C\overline{A}^t\overline{B}),\\
	y &= x * \overline{K},
\end{align}
where $*$ is convolution operation, and $t+1$ is the convolution kernel size. The recurrence mode is usually used for efficient autoregressive inference, while the convolution mode is used for efficient parallelizable training.

\subsection{Hidden State Degradation}
Here we analyze the hidden information flow from shallow layers to deep layers. In the following, we use the superscript ``$l$'' to represent the $l$-th block.

\begin{equation}\label{eq:hidden-analysis}
	\begin{aligned}
		h_t^l =& \overline{A}h_{t-1}^l + \overline{B}x_t^l\\
		=& \overline{A}h_{t-1}^l + \overline{B}\Theta(y_t^{l-1})\\
		=& \overline{A}h_{t-1}^l + \overline{B}\Theta(Ch_t^{l-1})\\
		=& \overline{A}h_{t-1}^l + \overline{B}\Theta(C\overline{A}h_{t-1}^{l-1} + C\overline{B}\Theta(Ch_t^{l-2}))\\
		=& \overline{A}h_{t-1}^l + \overline{B}\Theta(C\overline{A}h_{t-1}^{l-1} + \cdots  \\ & + C\overline{B}\Theta(C\overline{A}h_{t-1}^{l-m+1} + C\overline{B}\Theta(Ch_t^{l-m}\underbrace{))\cdots)}_{m},\\
	\end{aligned}
\end{equation}
where $\Theta(\cdot)$ is the transformations from the last output to the input of SSM module, such as convolution and FFN. From Eq.~\ref{eq:hidden-analysis}, we can see that the transmission of hidden information from the $(l-m)$-th layer to the $l$-th layer requires passing through $m$ transformation blocks and $m$ BC matrix multiplications. Such a complex computational process can lead to significant information loss, meaning that attempting to retrieve certain information from the $(l-m)$-th layer at the $l$-th layer becomes very challenging and unclear.


\subsection{Dense Hidden Connection}
Through the above analysis, we have identified a crucial issue in SSM, which is the decay of important hidden states as the layer depth increases. Therefore, we propose a dense connection for hidden states to better preserve fine-grained information from shallow layers, enhancing the ability of deep layers to perceive the original textual information. For the $l$-th block, we densely connect the hidden states in its previous $m$ blocks. First, we collect the shallow hidden states and introduce a selective transition module $\phi$ to project them to the subspace of the target layer and select useful parts simultaneously:
\begin{equation}
	\mathcal{H}_t^l = [\phi(h_t^{l-1});\phi(h_t^{l-2});\cdots;\phi(h_t^{l-m})],
\end{equation}
Then, the intermediate hidden vectors are injected into the original hidden state of this layer:
\begin{equation}\label{eq:fuse}
	h{'}_t^l = Fuse(h_t^l, \mathcal{H}_t^l).
\end{equation}
The operation $Fuse()$ is the function to fuse the intermediate hidden vectors and the current hidden state. The SSMs with the proposed dense hidden connection is named as DenseSSM (Figure~\ref{Fig:dense_ssm}(a)). The DenseSSM scheme can be used in any SSM variant such as Mamba~\cite{mamba}. Compared to DenseNet~\cite{huang2017densely} for convolutional networks, the proposed DenseSSM densely connect the hidden states in SSMs, and the selective mechanism and fusion manner are more efficient for language modeling.

The above analysis is based on the recurrence mode, in the following we introduce the convolution mode of DenseSSM for efficient training. From Eq.~\ref{eq:recurrence1}, we have
\begin{equation}
    \begin{aligned}
    h_t^l &= \overline{A}h_{t-1}^l + \overline{B}x_t^l\\
    &= \overline{A}(\overline{A}h_{t-2}^l + \overline{B}x_{t-1}^l) + \overline{B}x_t^l\\
    &= \overline{A}^2h_{t-2}^l + \overline{A}\overline{B}x_{t-1}^l + \overline{B}x_t^l\\
    &= \overline{A}^th_0^l + \overline{A}^{t-1}\overline{B}x_{1}^l + \cdots + \overline{A}\overline{B}x_{t-1}^l + \overline{B}x_t^l\\
    &= \overline{A}^t\overline{B}x_0^l + \overline{A}^{t-1}\overline{B}x_{1}^l + \cdots + \overline{A}\overline{B}x_{t-1}^l + \overline{B}x_t^l.
    \end{aligned}
\end{equation}
This process can be conducted by a convolution on the input sequence $(x_0^l,x_1^l,\cdots,x_t^l)$:
\begin{equation}
    \begin{aligned}
    h_t^l &= \overline{A}^t\overline{B}x_0^l + \overline{A}^{t-1}\overline{B}x_{1}^l + \cdots + \overline{A}\overline{B}x_{t-1}^l + \overline{B}x_t^l\\
    &= (x_0^l,x_1^l,\cdots,x_t^l) * (\overline{B},\overline{A}\overline{B},\cdots,\overline{A}^t\overline{B}).
    \end{aligned}
\end{equation}
In the proposed DenseSSM, we enhance the hidden states by Eq.~\ref{eq:fuse} and then obtain the outputs of SSM:
\begin{equation}
    \begin{aligned}
    y_t^l &= Ch{'}_t^l\\
    &= CFuse((x_0^l,x_1^l,\cdots,x_t^l) * (\overline{B},\overline{A}\overline{B},\cdots,\overline{A}^t\overline{B}), \mathcal{H}_t^l).
    \end{aligned}
\end{equation}
As shown in Figure~\ref{Fig:dense_ssm}(b), DenseSSM can be trained in parallelizable convolution mode.

\begin{figure}[htp]
	\centering
	\includegraphics[width=1.0\linewidth]{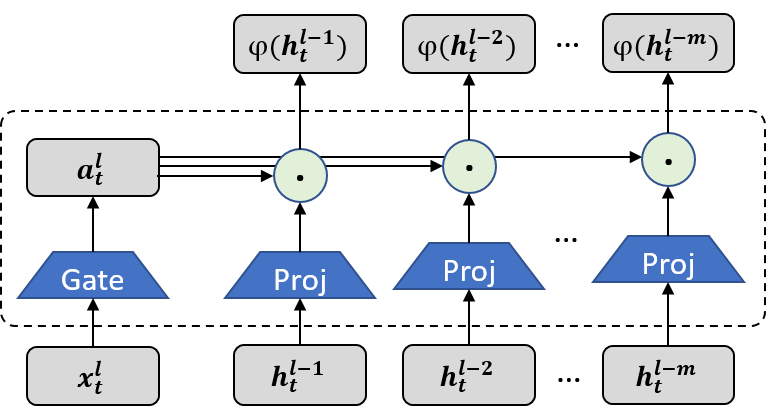}
	\vspace{-0.5em}
	\caption{Selective Transition Module.}
	\label{fig:select}
	\vspace{-0.5em}
\end{figure}

\paragraph{Selective Transition Module}
The selective transition module $\phi(\cdot)$ is to project inputs to the target subspace and select the useful part of hidden information simultaneously. We implement  the selective transition module with projection layer and gate selection mechanism, as shown in Figure~\ref{fig:select}. First, we project the hidden states in the previous $m$ SSM blocks to the same space:
\begin{equation}
    h{'}_t^{l-m} = Proj(h_t^{l-m}).
\end{equation}
Then we generate the gate weights based on the input $x_t^l$ and use them to select useful hidden states:
\begin{equation}
    \phi(h_t^{l-m}) = h{'}_t^{l-m}\odot Gate(x_t^l).
\end{equation}

Please note that the newly introduced modules must not compromise the training parallelizability and inference efficiency of the original SSM framework. Therefore, we maintain a simple and efficient implementation in practice. The projection layer is implemented using a linear transformation, while the gate module is implemented with a two-layer MLP with a SiLU activation~\cite{elfwing2018sigmoid}.

\paragraph{Hidden Fusion Module}
After the selective transition module, we obtain the selected hidden states from shallow layers, \emph{i.e.}, $\mathcal{H}_t^L = [\phi(h_t^1);\phi(h_t^2);\cdots;\phi(h_t^{L-1})]$. A hidden fusion module is utilized to integrate shallow hidden states with the current hidden states. Similarly, we keep the implementation simple for efficiency. We add the selected hidden states since they have been projected to the same space:
\begin{equation}
	h_t^L = Fuse(h_t^L, \mathcal{H}_t^L) = h_t^L + \sum_{i=1}^{m}h_t^{l-i}.
\end{equation}
Here, we provide a basic implementation, but of course, there are other implementation approaches such as concatenation and cross-attention. We will compare different implementation methods in later experiments.

\begin{figure}[htp]
	\centering
	\includegraphics[width=1.0\linewidth]{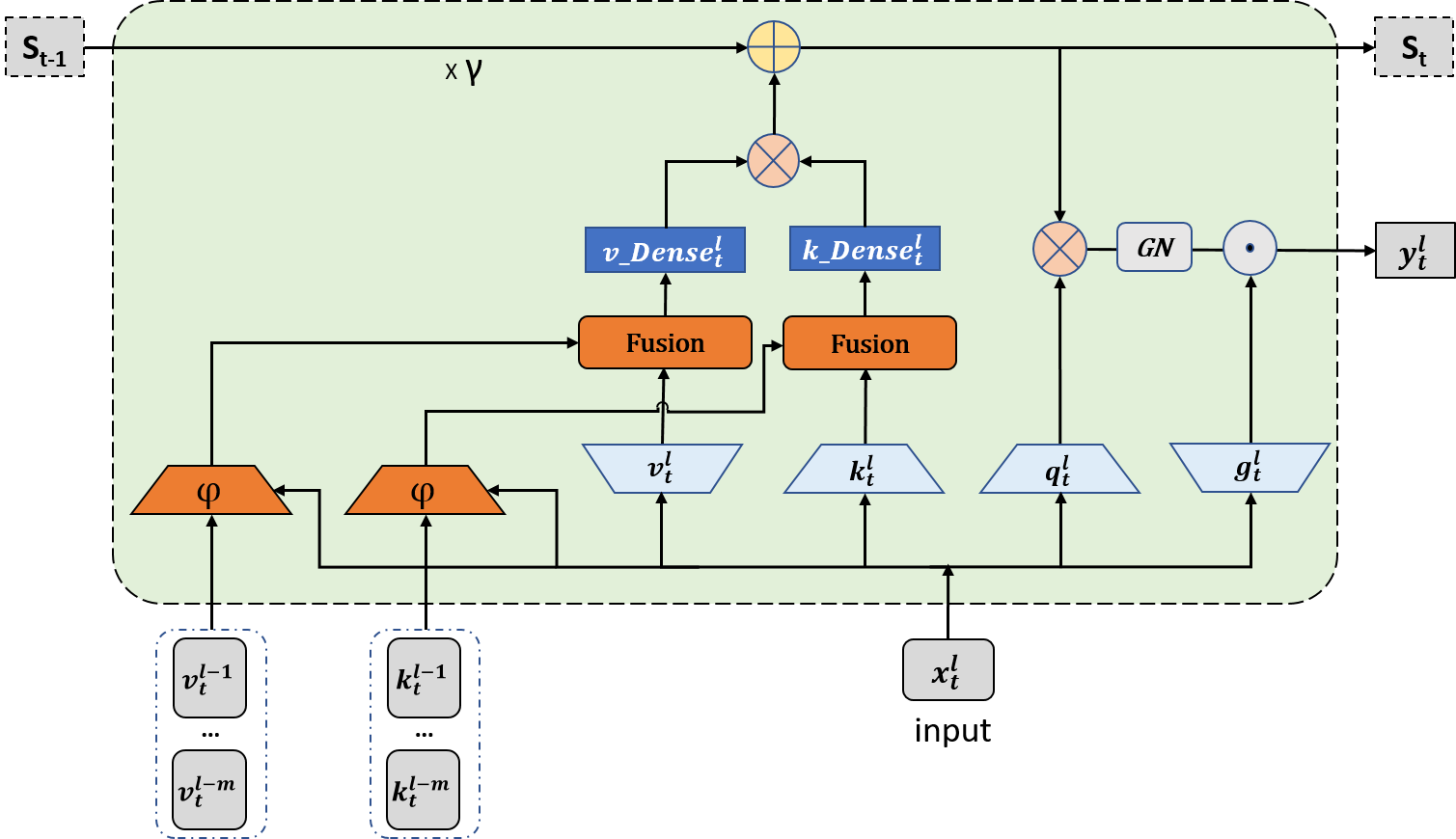}
	\vspace{-0.5em}
	\caption{DenseRetNet in autoregressive mode.}
	\vspace{-0.5em}
	\label{fig:denseretnet}
\end{figure}

\paragraph{Extension to RetNet}
RetNet~\cite{retnet} can be viewed as a kind of state space models which uses a variant of self-attention rather than convolution in Eq.~\ref{eq:conv}. Compared to the standard Transformer, RetNet is a RNN-style language model with fast inference and parallelized training. It utilizes linear attention to simplify the computation complexity of self-attention.
\begin{align}
    S_t &= \gamma S_{t-1} + k_t^Tv_t,\\
	y_t &= q_tS_t,
\end{align}
where $S_t$ is the recurrent state, and $0<\gamma<1$.
The dense KV connection for RetNet is performed as follows. The low-level keys and values are first concatenated:
\begin{align}
	\mathcal{K}_t^l &= [\phi(k_t^{l-1});\phi(k_t^{l-2});\cdots;\phi(k_t^{l-m})],\\
	\mathcal{V}_t^l &= [\phi(v_t^{l-1});\phi(v_t^{l-2});\cdots;\phi(v_t^{l-m})].
\end{align}
Then, the intermediate key (or value) vectors are injected into the original keys (or values) of this layer:
\begin{align}
	k{'}_t^L &= k_t^L + \sum_{i=1}^{m}k_t^{l-i},\\
	v{'}_t^L &= v_t^L + \sum_{i=1}^{m}v_t^{l-i}.
\end{align}
The RetNet equiped with the proposed dense key-value (KV) connections is named as DenseRetNet, as illustrated as shown in the figure~\ref{fig:denseretnet}. In addition, the paralleizable mode of DenseRetNet is formulated as follows:
\begin{equation}\label{eq:denseretnet-conv}
	y_t = q_t\sum_{i=1}^{t}\gamma^{t-i}k{'}_i^Tv{'}_i.
\end{equation}
Our DenseRetNet can be implemented in parallelizable mode as well, that is, can be trained in parallel on GPUs or NPUs.

\section{Experiments}
In this section, we conducted comprehensive experiments to validate the effectiveness of the proposed DenseSSM. The verification was carried out on different architectures, including RetNet and Mamba.

\subsection{Data and Experimental Settings}
\paragraph{Pretraining Data}
Following the common settings in~\cite{yang2023gated}, we trained all models from scratch utilizing a corpus comprising 56GB of raw data extracted from The Pile~\cite{gao2020pile}, a commonly used diverse and high-quality datasets. Excluding data from the DM$\_$Mathematics and Github subsets, we performed a random shuffle and sampled from all remaining corpus. The data was tokenized using the LLaMA tokenizer, which has a vocabulary size of 32,000 tokens. $<$bos$>$ token was used as the start-of-sequence marker. The resulting cached dataset contained a total of 15 billion tokens.

\paragraph{Evaluation Datasets}
In our experiment, we investigate models performance across a spectrum of downstream tasks, focusing on zero-shot and 4-shot learning capabilities. The tasks, presented in Table~\ref{tab:my-table-retnet} and ~\ref{tab:my-table-mamba}, encompass a range of datasets designed to test common-sense reasoning and question-answering, such as HellaSwag~\cite{zellers2019hellaswag}, BoolQ~\cite{clark2019boolq}, COPA~\cite{ponti2020xcopa}, PIQA~\cite{bisk2019piqa}, Winograd~\cite{muennighoff2022crosslingual}, Winogrande~\cite{sakaguchi2019winogrande}, StoryCloze~\cite{DBLP:journals/corr/abs-2112-10668}, OpenBookQA~\cite{OpenBookQA2018}, SciQ~\cite{Welbl2017CrowdsourcingMC}, ARC$\_$E(ARC-easy) and ARC$\_$C(ARC-challenge)~\cite{Clark2018ThinkYH}. 
Words Perplexity results of WikiText~\cite{merity2016pointer} and 
LAMBADA (LAMBADA$\_$OPENAI)~\cite{lambada} are also reported. All evaluations are executed using the LM evaluation harness~\cite{eval-harness}, ensuring a standardized approach to assessing the models' capabilities.

\subsection{Training Setup and Model's Architectures}
We selected the 350M and 1.3B model specifications to verify the validity of our proposed dense mechanism. All models were trained from scratch for one epoch on 15 billion tokens. The training batch size was set to 0.5 million tokens with a training length setting of 2048 tokens. AdamW \cite{loshchilov2019decoupled} optimizer was used for training, with a polynomial learning rate decay, and warm-up ratio is set to 1.5$\%$ of total training steps. Weight decay is set to 0.01, and gradient clipping is set to 1. 
We tailored the hyper-parameters of the model to ensure comparability with models of same scale. Additionally, we designed our Dense RetNet model to be fully comprised of GAU-like blocks, this  will be explicitly detailed in the subsequent paragraph.

\paragraph{Transformer-based language models}
We evaluate our proposed select dense mechanism against popular large language models like LLaMA~\cite{touvron2023llama} and OPT~\cite{zhang2022opt}, comparing with LLaMA for 350M size models and with OPT for 1.3B size models. Table~\ref{tab:hyper-llama-opt} reports their hyperparameters.

\begin{table}[h]
\centering
\begin{small}
\begin{tabular}{lccc}
\toprule
\bf Hyperparameters & \bf LLaMA 350M & \bf OPT 1.3B &  \\
\midrule
layers & 18 & 24 &  \\
hidden size & 1024 & 2048 &  \\
ffn size & 4096 & 8192 &  \\
heads & 8 & 32 &  \\
\midrule
learning rate & \multicolumn{3}{c}{$6\times10^{-4}$}  \\
Adam $\beta$ & \multicolumn{3}{c}{(0.9, 0.98)}  \\
dropout & 0.0 & 0.1  \\
\bottomrule
\\
\end{tabular}
\end{small}
\vskip -0.1in
\caption{Hyperparamters used for LLaMA and OPT models.
}
\label{tab:hyper-llama-opt}
\end{table}

\paragraph{Mamba}
As shwon in Table~\ref{tab:hyper-mamba}, since our tokenizer is smaller than the GPT-NeoX~\cite{black2022gptneox20b} tokenzier which Mamba~\cite{mamba} uses, we have added two additional layers to match the parameters. Besides this, we have adhered to Mamba's model structure and other training settings described in their paper. Specifically, we have set the learning rates to 3e-4 for the 360M model and 2e-4 for the 1.3M model, and we have not applied dropout in either case. The obtained new architecture is named as DenseMamba.

\begin{table}[h]
\centering
\begin{small}
\begin{tabular}{lccc}
\toprule
\bf DenseMamba \bf Hyperparameters & \bf360M & \bf1.3B &  \\
\midrule
n\underline{\hspace{0.5em}}layers & 50 & 50 &  \\
d\underline{\hspace{0.5em}}model & 1024 & 2048 &  \\
dense fusion layers & 4 & 4 &  \\

\midrule
learning rate & $3\times10^{-4}$ & $2\times10^{-4}$ & \\
Adam $\beta$ & \multicolumn{3}{c}{(0.9, 0.95)}  \\
dropout & \multicolumn{3}{c}{0.0} \\
\bottomrule
\\
\end{tabular}
\end{small}
\vskip -0.1in
\caption{Hyperparamters used for DenseMamba  models.
}
\label{tab:hyper-mamba}
\end{table}

\begin{table}[]
\centering
\begin{small}
\begin{tabular}{lccc}
\toprule
\bf DenseRetNet \bf Hyperparameters & \bf360M & \bf1.3B &  \\

\midrule
layers & 16 & 25 &  \\
hidden size & 1536 & 2560 &  \\
q $\&$ k size & 768 & 1280 &  \\
v $\&$ gate size & 3072 & 5120 &  \\
heads & 2 & 4 &  \\
dense fusion layers & 2 & 2 &  \\
\midrule
learning rate & \multicolumn{3}{c}{$6\times10^{-4}$} \\
Adam $\beta$ & \multicolumn{3}{c}{(0.9, 0.98)}  \\
dropout & \multicolumn{3}{c}{0.1} \\
\bottomrule
\\
\end{tabular}
\end{small}
\vskip -0.1in
\caption{Hyperparamters used for DenseRetNet models.
}
\label{tbl:hyper-denseretnet}
\end{table}

\begin{table*}[]
\resizebox{\textwidth}{!}{%
\setlength{\tabcolsep}{3pt}
\begin{tabular}{lcccccccccccccc}
\hline
\multicolumn{1}{l|}{\textbf{Models / Tasks}} & \multicolumn{1}{l}{\textbf{Wikitext}} & \multicolumn{1}{l|}{\textbf{LAMBADA}} & \multicolumn{1}{l}{\textbf{ARC\_C}} & \multicolumn{1}{l}{\textbf{ARC\_E}} & \multicolumn{1}{l}{\textbf{BoolQ}} & \multicolumn{1}{l}{\textbf{COPA}} & \multicolumn{1}{l}{\textbf{HellaSwag}} & \multicolumn{1}{l}{\textbf{PIQA}} & \multicolumn{1}{l}{\textbf{WinoGrande}} & \multicolumn{1}{l}{\textbf{StoryCloze}} & \multicolumn{1}{l}{\textbf{Winograd}} & \multicolumn{1}{l}{\textbf{OpenBookQA}} & \multicolumn{1}{l}{\textbf{SciQ}} & \multicolumn{1}{l}{\textbf{Avg.}} \\ \hline
\rowcolor[HTML]{EFEFEF} 
{\ul \textit{\textbf{Zero-Shot}}} & \multicolumn{1}{l}{\cellcolor[HTML]{EFEFEF}} & \multicolumn{1}{l}{\cellcolor[HTML]{EFEFEF}} & \multicolumn{1}{l}{\cellcolor[HTML]{EFEFEF}} & \multicolumn{1}{l}{\cellcolor[HTML]{EFEFEF}} & \multicolumn{1}{l}{\cellcolor[HTML]{EFEFEF}} & \multicolumn{1}{l}{\cellcolor[HTML]{EFEFEF}} & \multicolumn{1}{l}{\cellcolor[HTML]{EFEFEF}} & \multicolumn{1}{l}{\cellcolor[HTML]{EFEFEF}} & \multicolumn{1}{l}{\cellcolor[HTML]{EFEFEF}} & \multicolumn{1}{l}{\cellcolor[HTML]{EFEFEF}} & \multicolumn{1}{l}{\cellcolor[HTML]{EFEFEF}} & \multicolumn{1}{l}{\cellcolor[HTML]{EFEFEF}} & \multicolumn{1}{l}{\cellcolor[HTML]{EFEFEF}} & \multicolumn{1}{l}{\cellcolor[HTML]{EFEFEF}} \\
\multicolumn{1}{l|}{\textbf{LLaMa-350M}} & {\color[HTML]{333333} \textbf{26.79}} & \multicolumn{1}{c|}{{\color[HTML]{333333} 22.50}} & {\color[HTML]{333333} 22.95} & {\color[HTML]{333333} \textbf{46.13}} & {\color[HTML]{333333} \textbf{59.27}} & {\color[HTML]{333333} 64} & {\color[HTML]{333333} \textbf{33.19}} & {\color[HTML]{333333} \textbf{64.36}} & {\color[HTML]{333333} 49.09} & {\color[HTML]{333333} 57.64} & {\color[HTML]{333333} 62.02} & {\color[HTML]{333333} 29.6} & {\color[HTML]{333333} 75.3} & {\color[HTML]{333333} 51.23} \\
\multicolumn{1}{l|}{\textbf{RetNet-350M}} & {\color[HTML]{333333} 36.88} & \multicolumn{1}{c|}{{\color[HTML]{333333} 35.53}} & {\color[HTML]{333333} 21.25} & {\color[HTML]{333333} 40.99} & {\color[HTML]{333333} 48.35} & {\color[HTML]{333333} 61} & {\color[HTML]{333333} 29.86} & {\color[HTML]{333333} 62.30} & {\color[HTML]{333333} 51.07} & {\color[HTML]{333333} 55.59} & {\color[HTML]{333333} 59.05} & {\color[HTML]{333333} 28.4} & {\color[HTML]{333333} 75.8} & {\color[HTML]{333333} 48.51} \\
\multicolumn{1}{l|}{\textbf{DenseRetNet-350M}} & {\color[HTML]{333333} 31.35} & \multicolumn{1}{c|}{{\color[HTML]{333333} \textbf{19.92}}} & {\color[HTML]{333333} \textbf{23.72}} & {\color[HTML]{333333} 45.03} & {\color[HTML]{333333} 58.50} & {\color[HTML]{333333} \textbf{69}} & {\color[HTML]{333333} 32.31} & {\color[HTML]{333333} 64.04} & {\color[HTML]{333333} \textbf{52.09}} & {\color[HTML]{333333} \textbf{58.04}} & {\color[HTML]{333333} \textbf{60.82}} & {\color[HTML]{333333} \textbf{30.4}} & {\color[HTML]{333333} \textbf{76.6}} & {\color[HTML]{333333} \textbf{51.87}} \\ \hline
\rowcolor[HTML]{EFEFEF} 
{\ul \textit{\textbf{Four-Shot}}} & {\color[HTML]{333333} } & {\color[HTML]{333333} } & {\color[HTML]{333333} } & {\color[HTML]{333333} } & {\color[HTML]{333333} } & {\color[HTML]{333333} } & {\color[HTML]{333333} } & {\color[HTML]{333333} } & {\color[HTML]{333333} } & {\color[HTML]{333333} } & {\color[HTML]{333333} } & {\color[HTML]{333333} } & {\color[HTML]{333333} } & {\color[HTML]{333333} } \\
\multicolumn{1}{l|}{\textbf{LLaMa-350M}} & - & \multicolumn{1}{c|}{-} & {\color[HTML]{333333} 23.81} & {\color[HTML]{333333} \textbf{47.26}} & {\color[HTML]{333333} 53.00} & {\color[HTML]{333333} 65} & {\color[HTML]{333333} \textbf{33.71}} & {\color[HTML]{333333} \textbf{64.15}} & {\color[HTML]{333333} 51.14} & {\color[HTML]{333333} 57.38} & {\color[HTML]{333333} \textbf{64.25}} & {\color[HTML]{333333} \textbf{28.2}} & {\color[HTML]{333333} \textbf{81.2}} & {\color[HTML]{333333} \textbf{51.73}} \\
\multicolumn{1}{l|}{\textbf{RetNet-350M}} & - & \multicolumn{1}{c|}{-} & {\color[HTML]{333333} 23.04} & {\color[HTML]{333333} 40.91} & {\color[HTML]{333333} 50.37} & {\color[HTML]{333333} 63} & {\color[HTML]{333333} 29.49} & {\color[HTML]{333333} 62.08} & {\color[HTML]{333333} 51.78} & {\color[HTML]{333333} 55.66} & {\color[HTML]{333333} 59.61} & {\color[HTML]{333333} 27.4} & {\color[HTML]{333333} 77.4} & {\color[HTML]{333333} 49.16} \\
\multicolumn{1}{l|}{\textbf{DenseRetNet-350M}} & - & \multicolumn{1}{c|}{-} & {\color[HTML]{333333} \textbf{24.74}} & {\color[HTML]{333333} 45.66} & {\color[HTML]{333333} \textbf{54.89}} & {\color[HTML]{333333} \textbf{69}} & {\color[HTML]{333333} 32.14} & {\color[HTML]{333333} 63.70} & {\color[HTML]{333333} \textbf{52.01}} & {\color[HTML]{333333} \textbf{57.58}} & {\color[HTML]{333333} 59.23} & {\color[HTML]{333333} \textbf{28.2}} & {\color[HTML]{333333} 78.3} & {\color[HTML]{333333} 51.41} \\ \hline
\rowcolor[HTML]{EFEFEF} 
{\ul \textit{\textbf{Zero-Shot}}} & {\color[HTML]{333333} } & {\color[HTML]{333333} } & {\color[HTML]{333333} } & {\color[HTML]{333333} } & {\color[HTML]{333333} } & {\color[HTML]{333333} } & {\color[HTML]{333333} } & {\color[HTML]{333333} } & {\color[HTML]{333333} } & {\color[HTML]{333333} } & {\color[HTML]{333333} } & {\color[HTML]{333333} } & {\color[HTML]{333333} } & {\color[HTML]{333333} } \\
\multicolumn{1}{l|}{\textbf{OPT-1.3B}} & {\color[HTML]{333333} 22.04} & \multicolumn{1}{c|}{{\color[HTML]{333333} 13.79}} & {\color[HTML]{333333} \textbf{24.66}} & {\color[HTML]{333333} 48.65} & {\color[HTML]{333333} 58.07} & {\color[HTML]{333333} \textbf{63}} & {\color[HTML]{333333} 37.00} & {\color[HTML]{333333} 65.89} & {\color[HTML]{333333} \textbf{52.80}} & {\color[HTML]{333333} \textbf{61.02}} & {\color[HTML]{333333} 65.51} & {\color[HTML]{333333} 29.6} & {\color[HTML]{333333} 81.1} & {\color[HTML]{333333} 53.39} \\
\multicolumn{1}{l|}{\textbf{RetNet-1.3B}} & {\color[HTML]{333333} 27.90} & \multicolumn{1}{c|}{{\color[HTML]{333333} 23.41}} & {\color[HTML]{333333} 22.61} & {\color[HTML]{333333} 46.34} & {\color[HTML]{333333} 48.75} & {\color[HTML]{333333} 58} & {\color[HTML]{333333} 32.25} & {\color[HTML]{333333} 63.44} & {\color[HTML]{333333} 49.96} & {\color[HTML]{333333} 57.71} & {\color[HTML]{333333} 60.65} & {\color[HTML]{333333} 23.4} & {\color[HTML]{333333} 77.3} & {\color[HTML]{333333} 49.13} \\
\multicolumn{1}{l|}{\textbf{DenseRetNet-1.3B}} & {\color[HTML]{333333} \textbf{21.55}} & \multicolumn{1}{c|}{{\color[HTML]{333333} \textbf{10.88}}} & {\color[HTML]{333333} 24.49} & {\color[HTML]{333333} \textbf{50.88}} & {\color[HTML]{333333} \textbf{58.62}} & {\color[HTML]{333333} \textbf{63}} & {\color[HTML]{333333} \textbf{38.72}} & {\color[HTML]{333333} \textbf{67.25}} & {\color[HTML]{333333} 49.96} & {\color[HTML]{333333} 60.82} & {\color[HTML]{333333} \textbf{65.85}} & {\color[HTML]{333333} \textbf{31.8}} & {\color[HTML]{333333} 82.7} & {\color[HTML]{333333} \textbf{54.01}} \\ \hline
\rowcolor[HTML]{EFEFEF} 
{\ul \textit{\textbf{Four-Shot}}} & {\color[HTML]{333333} } & {\color[HTML]{333333} } & {\color[HTML]{333333} } & {\color[HTML]{333333} } & {\color[HTML]{333333} } & {\color[HTML]{333333} } & {\color[HTML]{333333} } & {\color[HTML]{333333} } & {\color[HTML]{333333} } & {\color[HTML]{333333} } & {\color[HTML]{333333} } & {\color[HTML]{333333} } & {\color[HTML]{333333} } & {\color[HTML]{333333} } \\
\multicolumn{1}{l|}{\textbf{OPT-1.3B}} & - & \multicolumn{1}{c|}{-} & {\color[HTML]{333333} \textbf{25.94}} & {\color[HTML]{333333} 50.46} & {\color[HTML]{333333} 52.35} & {\color[HTML]{333333} 63} & {\color[HTML]{333333} 36.97} & {\color[HTML]{333333} 64.64} & {\color[HTML]{333333} 52.33} & {\color[HTML]{333333} 60.09} & {\color[HTML]{333333} \textbf{66.58}} & {\color[HTML]{333333} \textbf{28.2}} & {\color[HTML]{333333} \textbf{89.4}} & {\color[HTML]{333333} 53.63} \\
\multicolumn{1}{l|}{\textbf{RetNet-1.3B}} & - & \multicolumn{1}{c|}{-} & {\color[HTML]{333333} 24.66} & {\color[HTML]{333333} 46.30} & {\color[HTML]{333333} 47.49} & {\color[HTML]{333333} \textbf{67}} & {\color[HTML]{333333} 31.96} & {\color[HTML]{333333} 63.22} & {\color[HTML]{333333} 52.09} & {\color[HTML]{333333} 57.51} & {\color[HTML]{333333} 61.42} & {\color[HTML]{333333} 26.6} & {\color[HTML]{333333} 80.3} & {\color[HTML]{333333} 50.78} \\
\multicolumn{1}{l|}{\textbf{DenseRetNet-1.3B}} & - & \multicolumn{1}{c|}{-} & {\color[HTML]{333333} 25.68} & {\color[HTML]{333333} \textbf{53.07}} & {\color[HTML]{333333} \textbf{56.3}} & {\color[HTML]{333333} \textbf{67}} & {\color[HTML]{333333} \textbf{38.56}} & {\color[HTML]{333333} \textbf{66.97}} & {\color[HTML]{333333} \textbf{53.59}} & {\color[HTML]{333333} \textbf{62.08}} & {\color[HTML]{333333} 65.12} & {\color[HTML]{333333} 27.8} & {\color[HTML]{333333} 86.7} & {\color[HTML]{333333} \textbf{54.81}} \\ \hline
\end{tabular}%
}
\caption{Benchmarking results of DenseRetNet are compared against the original RetNet~\cite{retnet} and Transformer-based models, specifically LLaMA-350M~\cite{touvron2023llama} and OPT-1.3B~\cite{zhang2022opt}. Our DenseRetNet architecture has lower perplexity and higher accuracy, effectively enhances the performance of Linear Attention, \emph{e.g.}, in RetNet, and surpasses the performance of Transformer-based Models.}
\vskip -0.1in
\label{tab:my-table-retnet}
\end{table*}
\paragraph{RetNet}
Model sizes and hyperparameters for our DenseRetNet is shown in Table~\ref{tbl:hyper-denseretnet}. We further utilize Gated Attention Unit (GAU) ~\cite{hua2022transformer} in our DenseRetNet. GAU combine Attention and FFN block into one, so a single block can perform both channel mixing and token mixing: $Y = (XW_u \odot A\hat{V})W_o$, where $A$ is attention weight cauculated though Eq.~\ref{eq:denseretnet-conv}. Also, multiple attention heads with different exponential decay rates are utilized to perform multi-scale decay instead of GAU's single-head strategy. In our experiments, we have observed that our architecture surpasses the RetNet structure with FFN layers in terms of training stability and performance.  The obtained new architecture is named as DenseRetNet.

\subsection{Main Results for DenseRetNet}
We evalute our models on both common corpus, and downstream tasks including common-sense reasoning and question-answering. Table~\ref{tab:my-table-retnet} presents the experimental results comparing DenseRetNet with LLaMA-350M~\cite{touvron2023llama}, OPT-1.3B~\cite{zhang2022opt} and RetNet~\cite{retnet}. Our DenseRetNet obtains lower perplexity on Wikitext and LAMBADA corpus and shows clear advantages in the downstream tasks in both 0-shot and few-shot settings. Especially, our model significantly improves the performance of RetNet, and achieves superior performance compared to the transformer large language models.

\subsection{Main Results for DenseMamba}
Table~\ref{tab:my-table-mamba} compares the performance of DenseMamba with LLaMA-350M~\cite{touvron2023llama}, OPT-1.3B~\cite{zhang2022opt}, and Mamba~\cite{mamba}. DenseMamba demonstrates superior perplexity and accuracy on the test set, outperforming Mamba and other Transformer-based models.

\subsection{Ablation Studies}
In this section, we conduct an ablation study to evaluate the impact of various design choices in our Selective Transition Module and Hidden Fusion Module. Perplexity results are presented for both in-domain evaluation sets and out-of-domain corpora~\cite{merity2016pointer}. For fair comparison, the baseline for all ablation studies is DenseRetNet-350M, with parameter adjustments to facilitate comparisons under similar computational constraints when necessary. We follow the default training settings outlined in Table \ref{tbl:hyper-denseretnet} for our models, except for the model trained on 1B tokens.

\paragraph{Ablations on Selective Transition Module}
The proposed selective transition module is to project the shallow hidden states to the same subspace and select the useful parts of them. The selective transition module can be implemented in different manners. 

Table~\ref{tbl:abl-select-transition} investigates the impact of various Projection and Select configurations. The experiment's other parameters were held constant: the number of dense layers(m) was set to 2, and the Fusion operation following the selective transition module was an "Add" operation. The findings suggest that the combination of Identity projection with MLP strikes an optimal balance between parameter count and performance.

\begin{table}[htbp]
    \centering
    \begin{small}
    \begin{tabular}{lc|cccc}
    \toprule
    \bf Projection & \bf Select & \bf \#Param & \bf In domain & \bf Wikitext\\
    
    \midrule
    None & None & 346M & 2.565 & 2.359  \\
    \midrule
    Identity & MLP &  353M & 2.546&2.348  \\
    Identity & Linear  & 357M & 2.572 & 2.369 \\
    Linear & MLP & 353M & 2.579 & 2.372   \\
    Linear & Linear & 356M & 2.582 & 2.378   \\
    \bottomrule
    \end{tabular}
    \end{small}
    \caption{In-domain evaluation cross-entropy loss and out-of-domain byte\_perplexity results for DenseRetNet-350M with various implementations of the selective transition module are presented. 
    }
    \label{tbl:abl-select-transition}
    \vskip -0.1in
\end{table}

\begin{table*}[]
\resizebox{\textwidth}{!}{%
\setlength{\tabcolsep}{3pt}
\begin{tabular}{lcccccccccccccc}
\hline
\multicolumn{1}{l|}{\textbf{Models / Tasks}} & \multicolumn{1}{l}{\textbf{Wikitext}} & \multicolumn{1}{l|}{\textbf{LAMBADA}} & \multicolumn{1}{l}{\textbf{ARC\_C}} & \multicolumn{1}{l}{\textbf{ARC\_E}} & \multicolumn{1}{l}{\textbf{BoolQ}} & \multicolumn{1}{l}{\textbf{COPA}} & \multicolumn{1}{l}{\textbf{HellaSwag}} & \multicolumn{1}{l}{\textbf{PIQA}} & \multicolumn{1}{l}{\textbf{WinoGrande}} & \multicolumn{1}{l}{\textbf{StoryCloze}} & \multicolumn{1}{l}{\textbf{Winograd}} & \multicolumn{1}{l}{\textbf{OpenBookQA}} & \multicolumn{1}{l}{\textbf{SciQ}} & \multicolumn{1}{l}{\textbf{Avg.}} \\ \hline
\rowcolor[HTML]{EFEFEF} 
{\ul \textit{\textbf{Zero-Shot}}} & \multicolumn{1}{l}{\cellcolor[HTML]{EFEFEF}} & \multicolumn{1}{l}{\cellcolor[HTML]{EFEFEF}} & \multicolumn{1}{l}{\cellcolor[HTML]{EFEFEF}} & \multicolumn{1}{l}{\cellcolor[HTML]{EFEFEF}} & \multicolumn{1}{l}{\cellcolor[HTML]{EFEFEF}} & \multicolumn{1}{l}{\cellcolor[HTML]{EFEFEF}} & \multicolumn{1}{l}{\cellcolor[HTML]{EFEFEF}} & \multicolumn{1}{l}{\cellcolor[HTML]{EFEFEF}} & \multicolumn{1}{l}{\cellcolor[HTML]{EFEFEF}} & \multicolumn{1}{l}{\cellcolor[HTML]{EFEFEF}} & \multicolumn{1}{l}{\cellcolor[HTML]{EFEFEF}} & \multicolumn{1}{l}{\cellcolor[HTML]{EFEFEF}} & \multicolumn{1}{l}{\cellcolor[HTML]{EFEFEF}} & \multicolumn{1}{l}{\cellcolor[HTML]{EFEFEF}} \\
\multicolumn{1}{l|}{\textbf{LlaMa-350M}} & 26.79 & \multicolumn{1}{c|}{22.50} & 22.95 & \textbf{46.13} & \textbf{59.27} & 64 & 33.19 & 64.36 & 49.09 & 57.64 & 62.02 & 29.6 & 75.3 & 51.23 \\
\multicolumn{1}{l|}{\textbf{Mamba-360M}} & 26.60 & \multicolumn{1}{c|}{17.55} & 23.98 & 45.83 & 55.78 & 61 & \textbf{34.89} & 64.31 & \textbf{52.88} & 58.90 & 62.92 & 29.2 & \textbf{79.8} & 51.77 \\
\multicolumn{1}{l|}{\textbf{DenseMamba-360M}} & \textbf{26.41} & \multicolumn{1}{c|}{\textbf{17.03}} & \textbf{24.32} & 46.0 & 59.20 & \textbf{66} & 34.68 & \textbf{64.80} & 51.14 & \textbf{59.03} & \textbf{63.23} & \textbf{29.8} & \textbf{79.8} & \textbf{52.55} \\ \hline
\rowcolor[HTML]{EFEFEF} 
{\ul \textit{\textbf{Four-Shot}}} &  &  &  &  &  &  &  &  &  &  &  &  &  &  \\
\multicolumn{1}{l|}{\textbf{LLaMa-350M}} & - & \multicolumn{1}{c|}{-} & 23.81 & \textbf{47.26} & 53.00 & 65 & 33.71 & 64.15 & 51.14 & 57.38 & 64.25 & 28.2 & \textbf{81.2} & 51.73 \\
\multicolumn{1}{l|}{\textbf{Mamba-360M}} & - & \multicolumn{1}{c|}{-} & \textbf{25.26} & 46.51 & 45.41 & 63 & 34.25 & \textbf{65.13} & \textbf{52.80} & \textbf{58.97} & 62.88 & \textbf{29.0} & 81.0 & 51.29 \\
\multicolumn{1}{l|}{\textbf{DenseMamba-360M}} & - & \multicolumn{1}{c|}{-} & 24.83 & 46.97 & \textbf{58.26} & \textbf{66} & \textbf{34.74} & 64.69 & 52.01 & 58.37 & \textbf{63.44} & 28.6 & 80.3 & \textbf{52.56} \\ \hline
\rowcolor[HTML]{EFEFEF} 
{\ul \textit{\textbf{Zero-Shot}}} &  &  &  &  &  &  &  &  &  &  &  &  &  &  \\
\multicolumn{1}{l|}{\textbf{OPT-1.3B}} & 22.04 & \multicolumn{1}{c|}{13.79} & 24.66 & 48.65 & 58.07 & 63 & 37.00 & 65.89 & \textbf{52.80} & 61.02 & 65.51 & 29.6 & \textbf{81.1} & 53.39 \\
\multicolumn{1}{l|}{\textbf{Mamba-1.3B}} & 21.79 & \multicolumn{1}{c|}{\textbf{12.46}} & \textbf{25.09} & 50.84 & 53.15 & \textbf{67} & 38.34 & 67.19 & 50.59 & 60.29 & 65.25 & 30.0 & 79.8 & 53.41 \\
\multicolumn{1}{l|}{\textbf{DenseMamba-1.3B}} & \textbf{21.39} & \multicolumn{1}{c|}{12.47} & \textbf{25.09} & \textbf{51.89} & \textbf{58.59} & \textbf{67} & \textbf{39.26} & \textbf{67.90} & 52.01 & \textbf{61.28} & \textbf{66.11} & \textbf{30.6} & 79.9 & \textbf{54.51} \\ \hline
\rowcolor[HTML]{EFEFEF} 
{\ul \textit{\textbf{Four-Shot}}} &  &  &  &  &  &  &  &  &  &  &  &  &  &  \\
\multicolumn{1}{l|}{\textbf{OPT-1.3B}} & - & \multicolumn{1}{c|}{-} & 25.94 & 50.46 & 52.35 & 63 & 36.97 & 64.64 & 52.33 & 60.09 & \textbf{66.58} & 28.2 & \textbf{89.4} & 53.63 \\
\multicolumn{1}{l|}{\textbf{Mamba-1.3B}} & - & \multicolumn{1}{c|}{-} & \textbf{26.96} & 52.69 & 49.56 & \textbf{69} & 39.25 & 66.27 & 52.96 & 61.15 & 66.06 & 30.4 & 82.3 & 54.24 \\
\multicolumn{1}{l|}{\textbf{DenseMamba-1.3B}} & - & \multicolumn{1}{c|}{-} & 26.54 & \textbf{52.99} & \textbf{58.59} & 67 & \textbf{39.26} & \textbf{67.08} & \textbf{53.67} & \textbf{61.48} & 65.89 & \textbf{31.0} & 82.1 & \textbf{55.05} \\ \hline
\end{tabular}%
}
\caption{Benchmarking results of DenseMamba are compared against LLaMA-350M~\cite{touvron2023llama}, OPT-1.3B~\cite{zhang2022opt}, and Mamba~\cite{mamba}. The results show that DenseMamba achieves a certain improvement achieving lower perplexity and higher accuracy on the test set compared to Mamba, and surpassing the capabilities of Transformer-based architectures.}
\label{tab:my-table-mamba}
\vskip -0.1in
\end{table*}

\paragraph{Ablations on Dense Layers}
In this experiment, we conducted an ablation analysis on the depth of fusion layers (denoted as ${m}$) . We employed a fusion strategy based on Identity projection and generate the gate using MLP. Our experimental results, as presented in Table \ref{tbl:abl-gate-numbers}, both two-layer (${m}$=2) and four-layer (${m}$=4) fusion architectures have performance gains. Considering the computational cost associated with training and inference, the two-layer fusion approach is considered to be more optimal.

In addition, we have explored the necessity of employing distinct gate generation layers for different dense layers. Our experimental results indicate that varying this configuration do not has a positive impact on the model's performance, which is beneficial for the development of lightweight dense connection architectures.

\begin{table}[htbp]
    \centering
    \begin{footnotesize}
    \begin{tabular}{cc|cccc}
    \toprule
        \bf Layers & \bf Diff. gates & \bf \#Param & \bf In domain & \bf Wikitext \\

    \midrule
    1 & \XSolidBrush &  353M & 2.570 & 2.363 \\
    2 & \XSolidBrush &  353M & 2.546 & 2.348 \\
    2 & \Checkmark  & 360M & 2.547  & 2.351\\
    4 & \XSolidBrush & 353M & 2.542 & 2.348\\
    4 & \Checkmark & 374M &  2.557 & 2.371 \\

    \bottomrule
    \end{tabular}
    \end{footnotesize}
    \caption{In-domain evaluation cross-entropy loss and out-of-domain byte\_perplexity results for DenseRetNet-350M with different dense layer numbers and different gate strategies. Diff. gates donates if different gate is applied to different dense features.
    }
    \label{tbl:abl-gate-numbers}
    \vskip -0.1in
\end{table}

\paragraph{Ablations on Hidden Fusion Module}
The hidden fusion module is to fuse the transited hidden states and the current hidden states. A popular way of feature fusion is by Concat followed by dimensionality reduction, which adds more number of parameters compared to our way. By fine-tuning the model structure, we compare it at the same magnitude, and Table~\ref{tbl:abl-add-concat} finds that our proposed lightweight Dense hidden connection achieves a better result.

\begin{table}[htbp]
    \centering
    \begin{small}
    \begin{tabular}{l|cccc}
    \toprule
    \bf Fusion & \bf \#Param & \bf In domain & \bf Wikitext  \\
    \midrule
    Concat & 354M & 2.551 & 2.370 \\
    Add  & 353M & 2.546 & 2.348  \\
    \bottomrule
    \end{tabular}
    \end{small}
    \caption{In-domain evaluation cross-entropy loss and out-of-domain byte\_perplexity of DenseRetNet-350M with different implementations of hidden fusion module.}
    \vskip -0.1in
    \label{tbl:abl-add-concat}

\end{table}

Another study investigates the impact of fusing dense features either every ${m}$ layers or at each individual layer. To maintain a consistent parameter count, we adjusted the dimensions of the MLP intermediate layer and trained the model with the full 15B tokens. The results in Table~\ref{tbl:abl-fuselocation} indicate that fusing at each layer more effectively facilitates information transfer from lower to higher layers.

\begin{table}[htbp]
    \centering
    \begin{small}
    \begin{tabular}{l|cccc}
    \toprule
    \bf Dense frequency & \bf \#Param & \bf In domain & \bf Wikitext  \\
    \midrule
    Every layer & 353M & 2.303 & 1.845 \\
    Every 2 layers & 353M & 2.331 & 1.866 \\
    Every 4 layers & 353M & 2.387 & 1.923 \\
    \bottomrule
    \end{tabular}
    \end{small}
    \caption{In-domain evaluation cross-entropy loss and out-of-domain byte-perplexity for DenseRetNet-350M with varying dense intervention frequency strategies.
    }
    \label{tbl:abl-fuselocation}
    \vskip -0.1in
\end{table}

\section{Conclusion}
In this paper, we propose a new DenseSSM framework for enhancing the hidden information flow cross different layers. The hidden states are crucial information storage units in the SSMs. Utilizing the hidden states from each layer more effectively would greatly benefit the fundamental capabilities of SSMs. Therefore, we propose to collect the hidden states from shallow layers and selectively fusing them into the hidden states of deeper layers to enhance the SSM's perception of low-level textual information. The proposed DenseSSM method does not affect the excellent characteristics of SSM, \emph{i.e.}, efficient autoregressive inference and efficient parallelizable training. We apply the DenseSSM method to widely used architectures like RetNet and Mamba, resulting in new architectures with stronger foundational language abilities and achieving higher accuracy in public benchmark evaluations.


\clearpage
\section{Impact Statements}
This paper presents work whose goal is to advance the field of Machine Learning. There are many potential societal consequences of our work, none which we feel must be specifically highlighted here.

\bibliography{example_paper}
\bibliographystyle{icml2024}

%
%

\end{document}